# RELATIONSHIP BETWEEN DIVERSITY AND PERFOMANCE OF MULTIPLE CLASSIFIERS FOR DECISION SUPPORT

**R. Musehane, F. Netshiongolwe, F.V. Nelwamondo*, L. Masisi and T. Marwala**

*School of Electrical & Information Engineering, University of the Witwatersrand, Private Bag 3, 2050, Johannesburg, South Africa*
**Graduate School of Arts and Sciences Harvard University GSAS Mail Center Child 412, 26 Everett Street Cambridge, Massachusetts, 02138 USA*

**Abstract:** The paper presents the investigation and implementation of the relationship between diversity and the performance of multiple classifiers on classification accuracy. The study is critical as to build classifiers that are strong and can generalize better. The parameters of the neural network within the committee were varied to induce diversity; hence structural diversity is the focus for this study. The hidden nodes and the activation function are the parameters that were varied. The diversity measures that were adopted from ecology such as Shannon and Simpson were used to quantify diversity. Genetic algorithm is used to find the optimal ensemble by using the accuracy as the cost function. The results observed shows that there is a relationship between structural diversity and accuracy. It is observed that the classification accuracy of an ensemble increases as the diversity increases. There was an increase of 3%-6% in the classification accuracy.

## 1. INTRODUCTION

Computational intelligence techniques have been used in many classification problems. The literature emphasises that a group of classifiers is better than one classifier [1-5]. This is because the decision that is made by a committee of classifiers is better than the decision made by one classifier. In this paper the committee of classifiers will be referred as an ensemble. The most popular way to gain confidence on the generalisation ability of an ensemble is by introducing diversity within the ensemble [1, 2, 5]. This has led to the development of measures of diversity and various aggregation schemes for combining classifiers. However, diversity is not clearly defined [6, 7]. Thus, a proper measure of diversity that will relate diversity to accuracy is to be adopted. Current methods commonly use the outcome of the individual classifiers of an ensemble to measure diversity. Hence an ensemble is considered diverse if classifiers within the ensemble produce different outcomes as opposed to having the same outcomes [1, 6, 7].

In this paper, as opposed to looking at the outcomes of the individual classifiers, ensemble diversity is viewed as the structural variation within classifiers that form an ensemble [1, 5]. Thus, diversity will be induced by changing structural parameters of a neural network [5]. The paper investigates the relationship between structural diversity within an ensemble and the prediction accuracy of the ensemble. It has been intuitively accepted that the classifiers to be combined should be diverse [8]. This is because it has been found meaningless to combine identical classifiers because no improvement can be achieved when combining them [8, 9]. Hence, measuring structural diversity and relating it to accuracy is crucial in order to build better learning machines. However, it is necessary to find the optimal size of an ensemble that gives better generalization. Therefore, a study on the size of the ensemble was done as to find the optimal size that can be used for the investigation. The methods for measuring structural diversity are to be devised and implemented. Moreover, the outcome diversity of structurally different classifiers is critical to be measured. This is because it is essential to show how correlated the outcomes of the structurally different classifiers is. Hence, the limitations of accuracy in the structural diversity are to be justified.

Different methods for creating diversity such as bagging and boosting have been explored [1, 3]. However, the aggregation methods are to be used to combine the ensemble predictions. Methods of voting and averaging have been found to be popular [9, 10] and hence are used in this study.

The paper first discusses the background in section 2. Analysis of the data used for this study is presented in section 3. The accuracy measure and structural measures of diversity used are discussed in section 4 and section 5. The methodologies used in investigating the effect of diversity on generalization are presented in section 6. The results and future work are then discussed in section 7.

## 2. BACKGROUND

### *2.1. Neural Networks*

Neural Networks (NN) are computational models that have the ability to learn and model linear and non-linear systems [11]. There are many types of neural networks but the most common neural network architecture is the multilayer perceptron (MLP) [11]. The neural network architecture that is used in this paper is a MLP network as shown in Figure 1. The MLP network has the input layer, the hidden layer and the output layer. An MLP network has parameters such as learning rate, number of hidden nodes and the activation function. These parameters can be varied to induce structural diversity [5]. The general equation of the output function of a MLP neural network is shown below (1).

$$y_k = f_{outer}(\sum_{j=1}^{M} w_{kj}^{(2)} f_{inner}(\sum_{i=1}^{N} w_{ij}^{(1)} x_j + w_{j0}^{(1)}) + w_{k0}^{(2)}) \quad (1)$$

where: $y_k$ is the output from the neural network, $f_{outer}$ is the output activation function that can be linear, softmax or logistic, $f_{inner}$ is the hidden layer tangential activation function. M is the number of the hidden units, N is the number of input units, $w_{kj}^{(2)}$ and $w_{ij}^{(1)}$ are the weights in the first and second layer moving from input i to hidden unit j, $w_{0j}^{(1)}$ and $w_{0k}^{(2)}$ are the biases for the unit j.

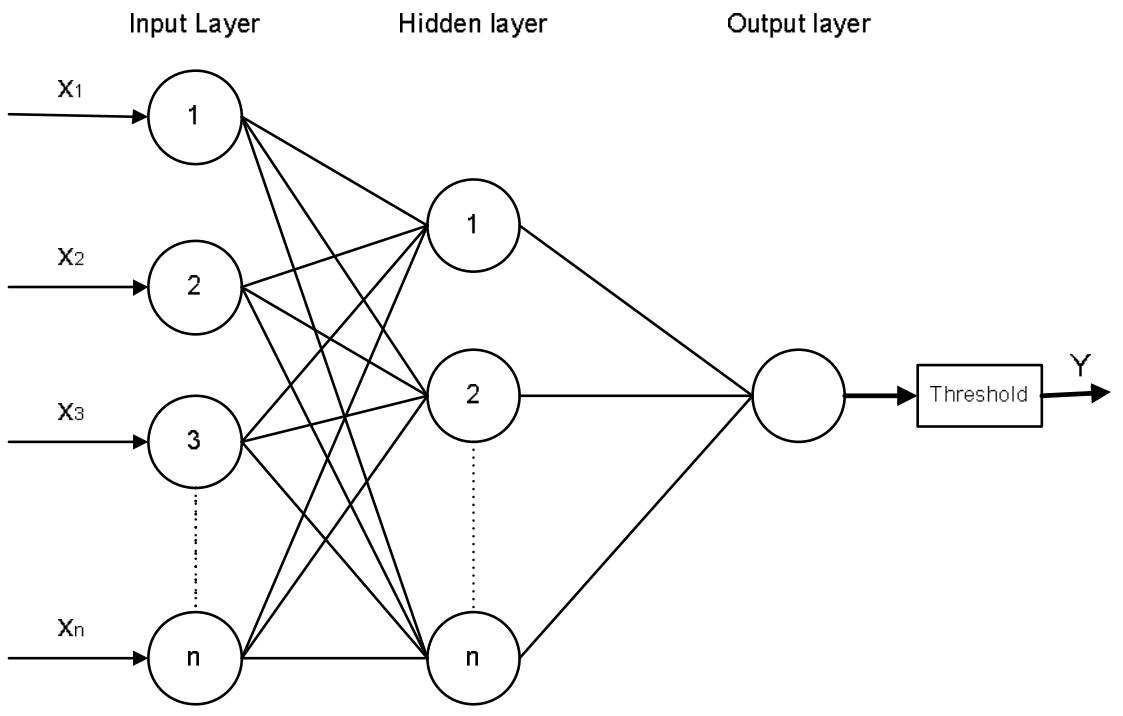

Figure 1: The MLP neural network architecture

The inputs into the neural network are the demographic data attributes from the HIV antenatal survey and the output is the HIV status of the individual where 0 represents negative and 1 represents positive. The weights of the NN are updated using a back propagation algorithm during the training stage [11].The threshold of 0.5 is used in order to achieve a zero or one solution from the neural network. This means that any value less than 0.5 is converted to 0 and any value more than 0.5 is converted to 1.

### 2.2. Genetic Algorithm

The genetic algorithms (GA) are computational models that are based on the evolution of biological population [2]. Potential solutions are encoded as the chromosomes of some individual. These individuals are initially generated randomly. The individuals are evaluated through the defined fitness function. Each preceding generation is populated by the fitness solution (members) of the previous generation and their offspring. The offsprings are created through crossover and mutation. The crossover process combines genetic information of two previous fittest solutions to create new offsprings. Mutation alters the genes of the individual to introduce more diversity into the population. In this way, the initial generated solution can be improved over time [2, 12].

## 3. DATA ANALYSIS

### 3.1. Data Collection

The dataset used for the study is from antenatal clinics in South Africa and it was collected by the department of health in 2001. The features in the data include the age, gravidity, parity, education, etc. The demographic data used in the study is shown in table 1 below. The province was provided as a string so it was converted to an integer from 1 to 9.

Table 1: The features from the survey

| | Variable | Type | Range |
|---|---|---|---|
| 1 | Age | integer | 13-50 |
| 2 | Education | integer | 0-13 |
| 3 | Parity | integer | 0-9 |
| 4 | Gravidity | integer | 1-12 |
| 5 | Province | integer | 1-9 |
| 6 | Age of father | integer | 14-60 |
| 7 | HIV status | binary | 0-1 |

The age is that of the mother visiting the clinic. Education represents the level of education the mother has and ranges from 1-13, where 1-12 corresponds to grade 1 to 12 and 13 represents tertiary education. Parity is the number of times the mother has given birth whilst gravidity is the number of times the mother has been pregnant. Both these quantities are important, as they show the reproductive activity as well as the reproductive health state of the women. The age of the father responsible for the current pregnancy is also given and the province entry corresponds to the geographic area where the mother comes from. The last feature is the HIV status of the mother where 0 represents a negative status whilst 1 represents a positive status.

### 3.2. Data Pre-Processing

The data preprocessing is necessary in order to eliminate impossible situations such as parity being greater than gravidity because it is not possible for the mother to give birth without falling pregnant. The pre-processing of the data resulted in a reduction of the dataset. To use the dataset for training, it needs to be normalized because some of the data variables with larger variances will influence the result more than others. This ensures that all variables can contribute to the final network weights of the prediction model [13]. Therefore, all the data is to be normalized between 0 and 1 using (2).

$$x_{norm} = \frac{x_i - x_{\min}}{x_{\max} - x_{\min}} \qquad (2)$$

where: $x_{\min}$ and $x_{\max}$ is the minimum and maximum value of the features of the data samples respectively.

The data were divided into three sets, the training, validation and testing data. This was done as to avoid over-fitting of the network. The neural networks are trained by 60% of the data, validated with 20% and tested with 20%.

## 4. MEASUREMENT OF ACCURACY

Regression problems mostly focus on using the mean square error between the actual outcome and the predicted outcome as a measure of how well neural networks are performing. In classification problems, the accuracy can be measured using the confusion matrix [14]. Analysis of the dataset that is being used showed that the data is biased towards the negative HIV status outcomes. Hence, the data was divided such that there is equal number of HIV positive and negative cases. The accuracy measure that is used in this study is given by (3).

$$Accuracy\ \% = \frac{TP + TN}{TP + TN + FP + FN} \times 100\% \qquad (3)$$

Where:
$TP =$ is the true positive -1 classified as a 1,
$TN =$ is the true negative - 0 classified as a 0,
$TN =$ is the false negative -1 classified as a 0,
$TP =$ is the false positive - 0 classified as a 1.

## 5. MEASUREMENT OF DIVERSITY

### 5.1. Shannon-Wiener Diversity Measure

Shannon entropy is a diversity measure that was adopted from ecology and information theory to understand ensemble diversity [15]. This measure is implemented to measure structural diversity. The Shannon-Wiener index is commonly used in information theory to quantify the uncertainty of the state [15, 16]. If the states are diverse one becomes uncertain of the outcome. It is also used in ecology to measure diversity of the species. Instead of biological species, the species are considered as the individual base classifiers. The Shannon diversity measure is given by (4).

$$D = -\frac{\sum_{i=1}^{M} \left(\frac{n_i}{N}\right) \ln\left(\frac{n_i}{N}\right)}{\log(N)} \qquad (4)$$

Where:

$n_i$ = number of neural networks that have the same structure
$N$ = total number of neural networks in an ensemble
$M$ = total number of different neural networks/species
$D$ = the diversity index

The diversity ranges from 0 to 1, where 0 indicates low diversity and 1 indicates highest diversity.

### 5.2. Simpson Diversity Measure

The other measure that was implemented is the Simpson diversity measure. This measure is also adopted from ecology to quantify diversity. It is quantified by (5).

$$D = \sum_{i=1}^{M} \frac{n_i(n_i - 1)}{N(N-1)} \qquad (5)$$

$n_i$ = number of neural networks that have the same structure
$N$ = total number of neural networks in an ensemble
$M$ = total number of different neural networks/species

The diversity index is given by $1 - D$. The diversity increases as the index increases. It ranges from 0 to 1 where 0 means there is no diversity and 1 indicate the highest diversity.

## 6. METHODOLOGY

### 6.1. Creation of Base Classifiers

Since the focus of the study is the structural diversity, the activation function, learning rate and the number of hidden nodes were varied as to induce diversity. However, varying all the parameters was found to be ineffective because the classifiers tend to generalize the same way. Therefore, only hidden nodes and activation function were varied for this investigation.

The classifiers are trained individually using the back propagation method; where the error is propagated back so as to adjust the weights accordingly. The data used for training, validation and testing are the HIV data. All the features of the input are fed to all the networks. The classifiers which have the training accuracy of 60% were accepted. The training accuracy between 60% and 63% was achieved. The hidden nodes were varied from 7 to 57 and the activation function between the logistics and the linear function was randomly varied. The classifiers were trained using quasi-Newton algorithm for 100 cycles at the same learning rate of 0.01.

### 6.2. Committee of Classifiers

The committee of classifiers improves efficiency and classification accuracy [17, 18]. This ensures that the results are based on the consensus decision of the base classifiers. The base classifiers operate concurrently during the classification and their outputs are integrated to obtain the final output [18]. The model for the committee of classifiers is shown in figure 2.

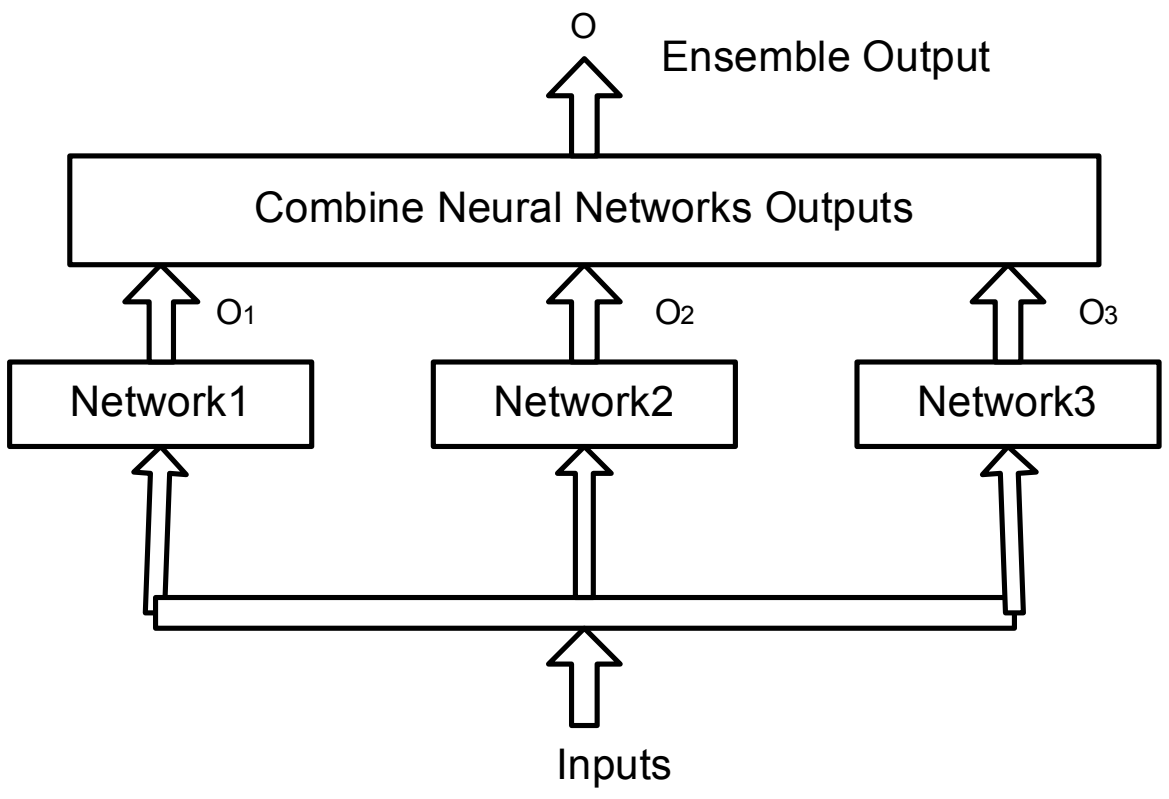

Figure 2: The classifier ensemble of neural networks

There are many aggregation methods that can be used to combine the outcomes of classifiers. These were explored in the preliminary report. The ensemble outcomes were all aggregated using simple majority voting. This was chosen because it is popular and easy to implement [9]. The outcomes of each individual from an ensemble are first converted to 0 or 1 using 0.5 as a threshold. The majority voting method chooses the prediction that is mostly predicted by different classifiers [19]. The other method that was implemented was averaging. All the outcomes from all the classifiers are taken and averaged.

### *6.3. Evaluation of Optimal Ensemble Size*

It is important to use the optimal size of an ensemble that results in better generalisation of the data [20]. The ensemble size is determined by the number of classifiers that belong to an ensemble. The created classifiers were used to carry out this experiment. The ensemble size was incremented by one from 1 to 50. However, the structure of the networks was made to be different by varying the hidden nodes as the ensemble size increases. Hence, the size of the network itself is increased as the number of classifiers in the ensemble increases [4]. Figure 3 below shows the results obtained.

It was however observed that the relationship between the size and accuracy of the ensemble depends on the accuracy of the individual classifiers that belong to the ensemble. Increasing the size of the neural network by increasing the hidden nodes tends to improve the classification accuracy as the number of the classifiers in an ensemble increases. However, an increase in size results in an increase in the prediction accuracy. Consequently, after the optimal size of 19 classifiers is reached, the accuracy tends to remain constant. Nevertheless, the size of 21 was found to be optimal since it produced the best accuracy. The results obtained are found to be concurrent with literature. Currently the optimal size of an ensemble is 25 [18, 20]. Therefore, an ensemble size of 21 is used for evaluating the relationship between diversity and performance of classifiers on HIV classification.

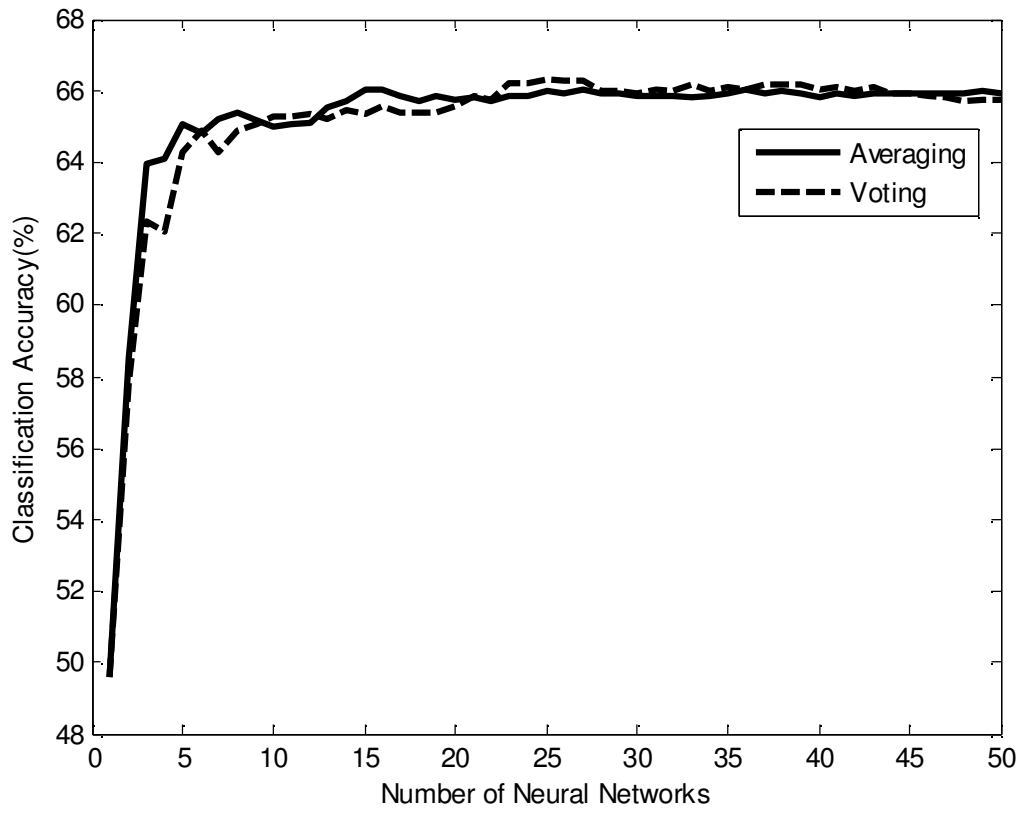

Figure 3: The ensemble size and classification accuracy

### *6.4. Evaluation of Outcome Diversity*

Currently, measuring the outcome diversity had been popular than measuring the structural diversity [6]. It was however necessary to measure the outcome diversity for this study. This is because it is essential to measure the degree of the agreement and disagreement on the outcomes of the ensemble. This experiment was useful for analysing the limitations on structural diversity results. The diversity measure such as Q statistics was used to measure diversity.

Q statistics evaluate the degree of similarity and dissimilarity in the outcomes of the classifiers within the ensemble [8]. The diversity index ranges from -1 to 1 where 0 indicates the highest diversity and 1 indicate lowest diversity [6]. For all 21 classifiers in an ensemble, each classifier is paired with every other classifier within the ensemble. The results from this study show that outcomes of the structurally diverse classifiers within the ensemble are highly correlated. This is indicated by a Q value which is closer to 1. The obtained Q value is from 0.88 to 0.91.

### *6.5. Evaluation of Structural Diversity*

The created classifiers were used to investigate the relationship between the diversity and accuracy. There were ten base classifiers or species that were selected from the created classifiers which are all structurally different based only on the hidden nodes and activation functions. These networks had different activation function and hidden nodes were varied from 10 to 55 in steps 5. The GA has the capabilities to search large spaces for a global optimal solution [5]. GA was therefore used to search for 21 classifiers from the 10 base classifiers using the accuracy as the fitness function. The fittest function is given by:

$$Fittest\ Function = -(T_{Acc} - Acc)^2 \qquad (6)$$

Where: $T_{Acc}$ is the targeted accuracy and $Acc$ is the obtained accuracy. The GA continues to search until the error between the targeted accuracy and the obtained accuracy is minimal. Firstly, it was necessary to optimize the accuracies that could be attained in order to minimize the computational cost. Thereafter, the attained accuracies were used in the second run as the target accuracy. The size of the neural network committee used is 21 classifiers which are formed from a combination of 10 unique base classifiers. Hence, each ensemble will have a repetition of certain classifiers. Once the ensemble of 21 classifiers produces the targeted diversity, the corresponding structural diversity is obtained using both Simpson and Shannon diversity measures given in (4) and (5). The algorithm implemented is shown in figure 4.

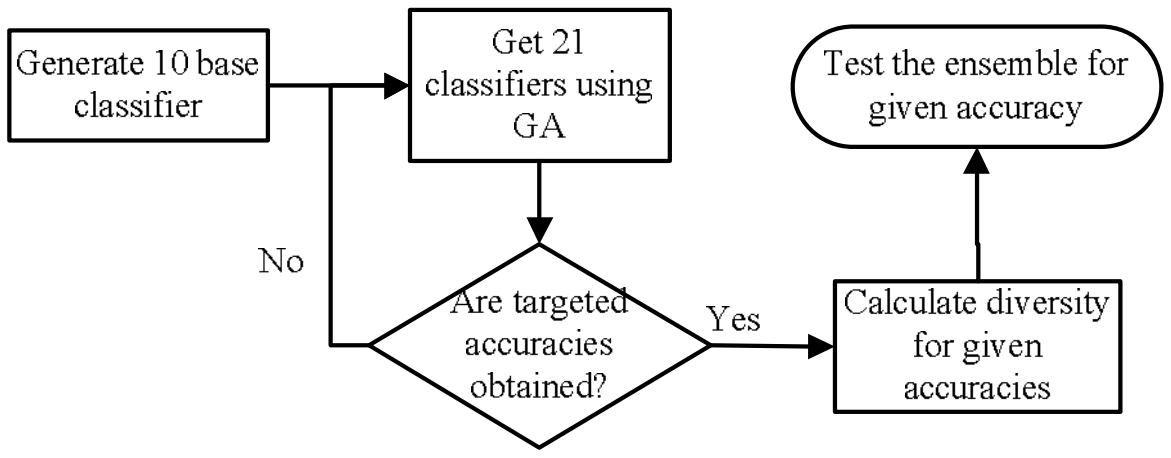

Figure 4: The algorithm used for evaluating diversity

## 7. RESULTS ANALYSIS

### *7.1. Structural Diversity Analysis*

In this study, diversity was induced by varying the parameters of the classifiers that form an ensemble [5, 16]. The investigation was done on an ensemble of 21 classifiers. Figure 5 shows the obtained results using the Shannon diversity measure. Figure 6 shows the results obtained using the Simpson diversity measure.

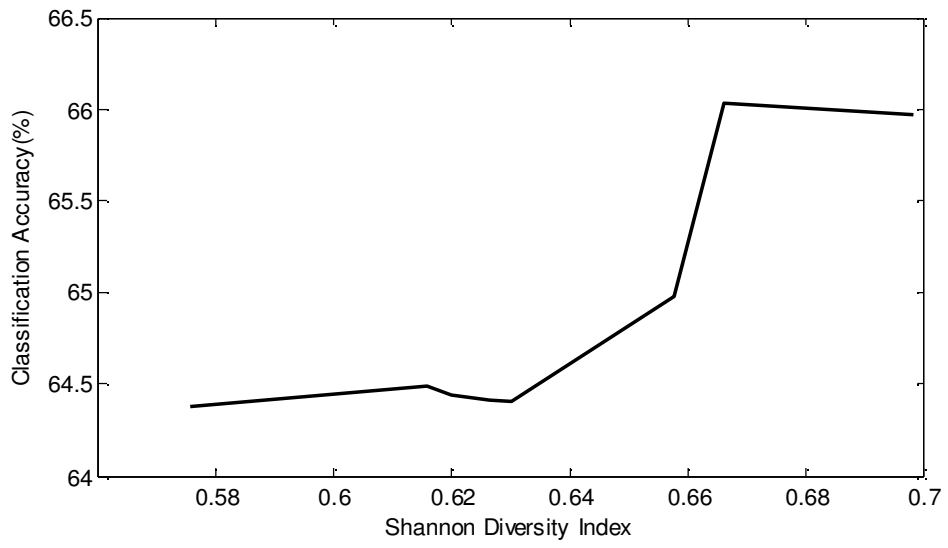

Figure 5: The evaluation of Shannon index with accuracy

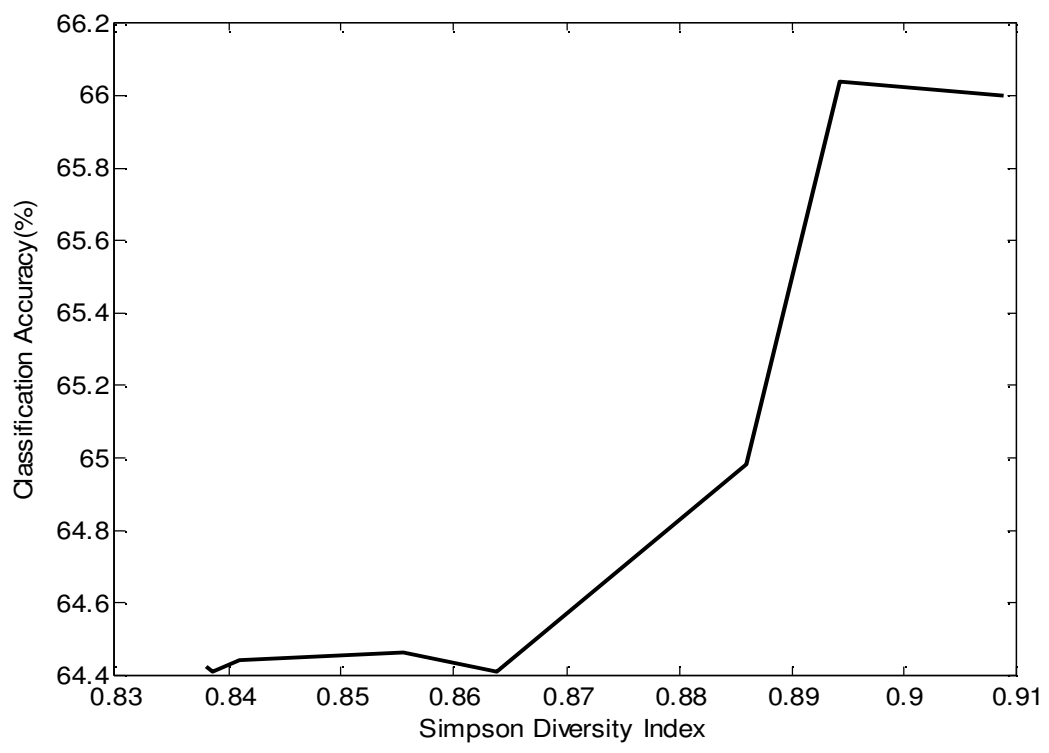

Figure 6: Evaluation of Simpson index with accuracy

The figures indicate that an increase in structural diversity results in an increase in accuracy which is in agreement with [16]. The experiment was done several times observing the relationship between diversity and accuracy using both Simpson and Shannon diversity measure. Therefore the results shown above are the average of ten different experiments that were performed. The results show that the two measures are concurrent. In the Shannon diversity measure, the GA was able to attain wide range of diversity whereas in the Simpson measure, the range is limited from 0.8 to 0.9. This was because the Shannon diversity index depends on the number of base classifiers whereas the Simpson's index depends on how evenly distributed the base classifiers are [15]. Shannon has shown that the more uncertain one is of the outcome, the more diverse an ensemble is.The results clearly show that structural variation of the parameters of the neural network (classifier) does have a relationship with classification accuracy As the structural diversity increased so did the accuracy.

### *7.2. Discussion and Recommendations*

It was however observed that the individual classifiers within the ensemble were highly correlated in the outcomes. This had affected the results because very low and high accuracies could not be attained. It is however recommended that a strategy of adding classifiers in an ensemble such that only classifiers that are uncorrelated are accepted in an ensemble can be adopted. The experiment focuses on training the classifiers using all the features of the data. It is however recommended that different networks can be fed different features of the data. This might ensure that the outcomes of classifiers are not highly correlated. Hence, a higher range of accuracy and diversity index can be attained.

During the training stage of the machine, the weights are normally randomly initialised. However, it has been found that different initial weights induce diversity within the ensemble [1]. The Shannon and Simpson diversity measures focuses on how structurally different the classifiers in an ensemble are. These measures do not consider diversity induced during initialisation of weights. Therefore, it is recommended that for future work, a better measure of structural diversity that incorporates the effect of weight initialisation should be developed.

## 8. CONCLUSION

The paper presented the relationship between structural diversity and generalization accuracy using Shannon and Simpson diversity measures to quantify diversity. The investigation is necessary as to build learning machines or committee of networks that can generalize better. The results have clearly shown that as the structural diversity index based on the measures used increases, the ensemble accuracy increases. Hence, the classifiers can be made structurally different in order to gain good classification accuracy. This has brought an increase of 3% to 6% in the classification accuracy. The method used to compute the results was found to be computationally expensive due to the use of GA. There is however limitations brought about by the individual classifiers producing similar outcomes even though they are structurally different. However, the use of measuring structural diversity in building good ensembles of classifiers is still to be explored.

## ACKNOWLEDGEMENT

The author would like to thank Fulufhelo Netshiongolwe for his cooperation and contribution during the project as a project partner. Professor Tshilidzi Marwala is thanked for supervising the project and additional thanks are extended to the postgraduate student Lesedi Masisi for his contribution during implementation of the project.